\pdfoutput=1
\documentclass[11pt]{article}

\usepackage{emnlp2021}

\usepackage{times}
\usepackage{latexsym}

\usepackage[T1]{fontenc}
\usepackage[utf8]{inputenc}

\usepackage{microtype}

\usepackage{stfloats}
\usepackage{amsmath}
\usepackage{amsbsy}
\usepackage{amssymb}
\usepackage{graphicx}
\usepackage[export]{adjustbox}
\usepackage{subcaption}
\usepackage{footnote}
\usepackage{multirow}
\makesavenoteenv{tabular}
\makesavenoteenv{table}
\usepackage{url}
\usepackage{tikz}
\usetikzlibrary{fit,positioning}
\usepackage{booktabs}
\usepackage{tabularx}
\usepackage{makecell}
\usepackage[ruled,vlined]{algorithm2e}

\usepackage[english]{babel} % added by djamé

\DeclareMathOperator*{\KL}{KL}
\DeclareMathOperator*{\ELBo}{ELBo}

\usepackage{microtype}

\usepackage{xcolor}

\expandafter\def\expandafter\UrlBreaks\expandafter{\UrlBreaks%  save the current one
  \do\a\do\b\do\c\do\d\do\e\do\f\do\g\do\h\do\i\do\j%
  \do\k\do\l\do\m\do\n\do\o\do\p\do\q\do\r\do\s\do\t%
  \do\u\do\v\do\w\do\x\do\y\do\z\do\A\do\B\do\C\do\D%
  \do\E\do\F\do\G\do\H\do\I\do\J\do\K\do\L\do\M\do\N%
  \do\O\do\P\do\Q\do\R\do\S\do\T\do\U\do\V\do\W\do\X%
  \do\Y\do\Z}

\usepackage{soulutf8}
\usepackage[normalem]{ulem}

\usepackage{enumitem}
\setlist{nolistsep}
\usepackage{microtype}

\iftrue % set to iftrue for camera ready version, iffalse to display draft comments
  \usepackage[final]{proofing}
  \renewcommand\hl[1]{{#1}}  %% to remove the highlith
  {\draftnote{\red{#2}}}
  \newcommand\redHL[1]{}
  \newcommand\todo[1]{}
  \newcommand{\Djame}[1]{}

\newcommand{\gfcmt}[1]{}

\newcommand{\jlcmt}[1]{}

\newcommand{\dscmt}[1]{}

\else  % the stuff below is for the draft mode (hl, comments and so)
\usepackage[draft]{proofing}

\newcommand{\gfcmt}[1]{\textcolor{orange}{#1}}

\newcommand{\jlcmt}[1]{\textcolor{red}{#1}}

\newcommand{\dscmt}[1]{\textcolor{brown}{#1}}

\newcommand{\Djame}[1]{
\textbf{\textcolor{red}{\hl{Djame: #1}}}
}

\newcommand\red[1]{{\textbf{\textcolor{red}{#1}}}}

\let\oldred\red
\renewcommand\red[1]{{\bf \oldred{{#1}}}}

 \newcommand\redHL[1]{\red{\hl{#1}}}
\let\olddraftnote\draftnote
\renewcommand\draftnote[1]{\olddraftnote{\red{#1}}}

\fi

\title{Challenging the Semi-Supervised VAE Framework for Text Classification}

\author{Ghazi Felhi,\hspace{5px} Joseph Le Roux\\
  LIPN\\ Université Sorbonne Paris Nord - CNRS UMR 7030 \\
  F-93430, Villetaneuse, France \\
  \texttt{\{felhi, leroux\}@lipn.fr} \\\And 
  Djamé Seddah \\
  INRIA Paris \\ Paris, France \\
 \hspace{5px} \texttt{djame.seddah@inria.fr} \\}
\begin{document}
\maketitle
\begin{abstract}
  Semi-Supervised Variational Autoencoders (SSVAEs) %\cite{Kingma2014a} 
  are widely used models for data efficient learning.
In this paper, we question the adequacy of the standard design of sequence SSVAEs for the task of text classification as we exhibit two sources of overcomplexity for which we provide simplifications.

  These simplifications to SSVAEs preserve their theoretical soundness while providing a number of practical advantages in the semi-supervised setup where the result of training is a text classifier.
  These simplifications are the removal of \textit{(i)} the Kullback-Liebler divergence from its objective and \textit{(ii)} the fully unobserved latent variable from its probabilistic model.
  These changes relieve users from choosing a prior for their latent variables, make the model smaller and faster, and allow for a better flow of information into the latent variables.

  We compare the simplified versions to standard SSVAEs on 4 text classification tasks.
  On top of the above-mentioned simplification, experiments show a speed-up of 26\%, while keeping equivalent classification scores.
  The code to reproduce our experiments is public\footnote{\href{https://github.com/ghazi-f/Challenging-SSVAEs}{https://github.com/ghazi-f/Challenging-SSVAEs}}.

\end{abstract}

\section{Introduction}

Obtaining labeled data to train NLP systems is a process that has often proven to be costly and time-consuming, and this is still largely the case \cite{martinez-alonso-etal-2016-noisy,Seddah2020BuildingHell}.
Consequently, semi-supervised approaches are appealing to improve performance while alleviating dependence on annotations.
To that end, Variational Autoencoders (VAEs) \cite{Kingma2014Auto-encodingBayes} have been adapted to semi-supervised learning \cite{Kingma2014a}, and subsequently applied to several NLP tasks \cite{Chen2018VariationalLearning,Corro2019DifferentiableAutoencoder,Gururangan2020VariationalClassification}.

A notable difference between the generative model case from where VAEs originate, and the semi-supervised case is that only the decoder (generator) of the VAE is kept after training in the first case, while in the second, it is the encoder (classifier) that we keep. This difference, as well as the autoregressive nature of text generators has not sufficiently been taken into account in the adaptation of VAEs to semi-supervised text classification.
In this work, we show that some components can be ablated from the long used semi-supervised VAEs (SSVAEs) when only aiming for text classification. These ablations simplify SSVAEs and offer several practical advantages while preserving their performance and theoretical soundness.

The usage of unlabeled data through SSVAEs is often described as a \emph{regularization}
%(or \emph{over-regularization})
on representations \cite{Chen2018VariationalLearning,Wolf-Sonkin2018AInflection,Yacoby2020FailureTasks}.
More specifically, SSVAEs add to the supervised learning signal, a conditional generation learning signal that is used to train on unlabeled samples.
From this observation, we study two changes to the standard SSVAE framework.
The first simplification we study is the removal of a term from the objective of SSVAEs: the Kullback-Leibler term.
This encourages the flow of information into latent variables, frees the users from choosing priors for their latent variables, and is harmless to the  theoretical soundness of the semi-supervised framework.
The second simplification we study is made to account for the autoregressive nature of text generators. In the general case, input samples in SSVAEs are described with two latent variables: a partially-observed latent variable, which is also used to infer the label for the supervised learning task, and an unobserved latent variable, which describes the rest of the variability in the data.
However, autoregressive text generators are powerful enough to converge without the need for latent variables. Therefore, removing the unobserved latent variable is the second change we study in SSVAEs. The above modifications can be found in some rare works throughout the literature, \textit{e.g.} \cite{Corro2019DifferentiableAutoencoder}.  We, however, aim to provide justification for these changes beyond the empirical gains that they exhibit for some tasks.

Our experiments on four text classification datasets show no harm to the empirical classification performance of SSVAE in applying the simplifications above. Additionally, we show that removing the unobserved latent variable leads to a significant speed-up.

To \draftreplace{sum up}{summarize} our contribution, we justify two simplifications to the standard SSVAE framework, explain the practical advantage of applying these modifications, and provide empirical results showing that they speed up the training process while causing no harm to the classification performance.

\section{Background}
\subsection{Variational Autoencoders}
Variational Autoencoders \cite{Kingma2019AnAutoencoders} are a class of generative models that combine Variational Inference with Deep Learning modules to train a generative model.
For a latent variable $z$, and an observed variable $x$, the generative model $p_\theta$ consists of a prior $p_\theta(z)$ and a decoder $p_\theta(x|z)$. VAEs also include an approximate posterior (also called the encoder) $q_\phi(z|x)$. Both are used during training to maximize an objective called the Evidence Lower Bound (ELBo), a lower-bound of the log-likelihood:

 \begin{align}
\log p_\theta(x) \geq& \nonumber\\
\mathbb{E}_{z\sim q_\phi(z|x)}&\left[\log p_\theta(x|z)\right] %\\&&
-\KL\left[q_\phi(z|x);p_\theta(z)\right]\nonumber\\
&=\ELBo(x; z) \label{ELBO}
 \end{align}

Throughout the paper, we will continue to use this $\ELBo(.;.)$ operator, with the observed variable(s) as a first argument, and the latent variable(s) as a second argument. In the original VAE framework, after training, the encoder $q_\phi$ is discarded and only the generative model (the prior and the decoder) are kept.

\subsection{Semi-Supervised VAEs}
The idea of using the VAE encoder as a classifier for semi-supervised learning has first been explored in \cite{Kingma2014a}. 
Besides the usual unobserved latent variable $z$, the semi-supervised VAE framework also uses a partially-observed latent variable $y$.
The encoder $q_\phi(y|x)$ serves both as the inference module for the supervised task, and as an approximate posterior (and encoder) for the $y$ variable in the VAE framework. 

Consider a set of labeled examples $L=\{(x_1, y_1), ..., (x_{|L|}, y_{|L|})\}$, and a set of unlabeled examples $U=\{x'_1, ..., x'_{|U|}\}$. For the set $L$, $q_\phi(y|x)$ is trained \textit{i)} with the usual supervised objective (typically, a cross-entropy objective for a classification task) \textit{ii)} with an ELBo that considers $x$ and $y$ to be \emph{observed}, and $z$ to be a \emph{latent} variable. A weight $\alpha$ is used on the supervised objective to control its balance with ELBo. For the set $U$, $q_\phi(y|x)$ is only trained as part of the VAE model with an ELBO where $y$ is used, this time, as a \emph{latent} variable like $z$. Formally, the training objective $ \mathcal{J}^\alpha$ of a SSVAE is as follows:

 \begin{align}
  \mathcal{J}^\alpha &= \sum_{(x, y) \in L}\Bigr( \ELBo((x, y);z)+ \alpha \hspace{1mm} \log q_\phi(y|x)\Bigl)\nonumber\\
  &+ \sum_{x \in U} \ELBo(x;(y, z)) \label{JALPHA}
 \end{align}
 
\section{Simplifying SSVAEs for Text Classification}
\label{ABLATING}
The simplifications we propose stem from the analysis of an alternative form under which ELBO can be written (Eq. 2.8 in \citealp{Kingma2019AnAutoencoders}). 
Although it is valid for any arguments of $\ELBo(.;.)$, we display it here for an observed variable $x$, and the couple of latent variables $(y,z)$:\
\begin{align}
\ELBo(&x; (y, z))= \nonumber\\
\log p_\theta&(x) - \KL[q_\phi(y, z|x)||p_\theta(y, z|x)] \label{AltELBO}
\end{align}
For the case of SSVAEs, this form provides a clear reading of the additional effect of ELBo on the learning process: \textit{i)} maximizing the log-likelihood of the generative model $p_\theta(x)$, \textit{ii)} bringing the parameters of the inference model $q_\phi(y, z|x)$ closer to the posterior of the generative model $p_\theta(y, z|x)$.
Since $p_\theta(y, z|x)$ is the distribution of the latent variables expected by the generative model $p_\theta$ for it to be able to generate $x$,
%Therefore, in addition to the supervised learning signal,
%it can be seen through this perspective
we can conclude that ELBo trains \emph{both} latent variables for conditional generation on the unsupervised dataset $U$. %When $\ELBo((x, y), z)$ is written under the same form, it can also be seen that only $z$ is trained on conditional generation for the labeled examples in $L$.

\subsection{Dropping the Unobserved Latent Variable}
Building on observations from equation \ref{AltELBO}, we question the usefulness of training both latent variables for conditional generation when semi-supervised learning only aims for an improvement on the inference of the partially-observed latent variable $y$.

%VAEs' generative capabilities have first been exhibited on image datasets \cite{Kingma2014Auto-encodingBayes}. For such datasets, the reconstruction loss (first term in Eq. \ref{ELBO}) is an L2 loss (or an L1 loss). This stems from the fact that $p_\theta(x|z)$ is modeled by a Gaussian (or resp. a Laplace) distribution. Given a single sample $z$, the blur modeled by such distributions cannot (in the general case) model an entire image dataset. The reader may refer to \citet{Zhao2017TowardsModels} for a full discussion of this issue. As a consequence, it is necessary for such datasets to incorporate a latent variable $z$ besides the partially-observed variable $y$ to model the characteristics that $y$ does not describe.

For the case of language generation, the sequence of discrete symbols in each sample is often modeled by an autoregressive distribution $p_\theta(x|y, z) = \prod_i p_\theta(x_i|y, z, x_{<i})$ where $x_i$ is the $i^{th}$ symbol in the sequence, and $x_{<i}$ are the symbols preceding $x_i$. Such a distribution is able to generate realistic samples when trained on a target text corpus, so much that text VAEs are plagued with a problem known as \emph{posterior collapse} \cite{Bowman2016GeneratingSpace} where the latent variable is ignored by the generative model. %It is also worth noting that current state of the art language modeling techniques rely on auto-regressive models that do not use latent variables \cite{Radford2018LanguageLearners, Melis2019MogrifierLSTM, Brown2020LanguageLearners}.
%Given the above $p_\theta(x)=\int_y p_\theta(x_0|y)\prod_i p_\theta(x_i|y, x_{<i}) dy$, an autoregressive language model that incorporates only $y$ as a latent variable is an expressive enough generative model to provide quality learning signal for $y$'s training on conditional generation (Eq. \ref{AltELBO}). 
We therefore propose to keep only $y$ and to drop $z$ from the model avoiding its presence in
the Kullback-Leibler divergence in Equation \ref{AltELBO} and saving some
parameters.\footnote{ In this case, one may be tempted to drop the VAE
  framework entirely and resort to other learning algorithms such as EM or
  direct likelihood maximization. Although possible in theory, this would  disconnect $q_\phi$ from the generator's training, and thus discard the benefit from using unlabeled data.
}
% There is a caveat regarding this modification: Since using only $y$ often makes integration over the latent variables possible, one may be tempted to optimize the exact log-likelihood instead of ELBo. This should not be done as it would remove the second term from Eq. \ref{AltELBO}, decoupling the learning processes of the generative model from that of the inference model, and therefore discarding the benefit provided by semi-supervised learning with VAEs. Nevertheless, keeping only $y$ still often enables calculating exactly the reconstruction term (the expectation in Eq. \ref{ELBO}) which presents no harm to the learning process.

\subsection{Dropping the Kullback-Leibler Term}
Previous work on VAE-based language models showed that the KL divergence in Eq.~\ref{ELBO} sometimes discourages the model from using latent variables and makes them useless in practice \cite{Bowman2016GeneratingSpace, DBLP:journals/corr/ZhaoSE17a, Chen2018c}.

An interesting result from \citet{DBLP:journals/corr/ZhaoSE17a} is that ELBo without KL divergence (\emph{KL-free}) is still a theoretically sound objective for generative modeling with VAEs.
The difference between the generative model resulting from a regular ELBo and a KL-free ELBo is the prior of the model.
A KL-free ELBo results in a generative model that uses as a prior $q_\phi(z)=\int_z q_\phi(z|x)p_{data}(x)dx$.
This prior is intractable which makes the resulting model impractical for generation, but causes no problem for semi-supervised VAEs.
We therefore propose, as a second change to the standard SSVAE framework, the removal of the KL-divergence in Eq.~\ref{ELBO}.

Note that in this case, the network formulates its own prior instead of requiring the user to choose it.
That is a significant advantage since the choice of a good prior is difficult: it must model adequately the \emph{default} behavior of the latent variables, and requires a closed form for the KL-divergence in Eq.~\ref{ELBO} to stabilize training.

% Removing the Kullback-Liebler term causes the network to train its own prior instead of using one that is chosen by the practitioner. Priors for partially-observed latent variables are delicate to choose as it is highly preferred for them to \textit{i)} yield a closed form or the KL-divergence in Eq. \ref{ELBO} to stabilize training, \textit{ii)} realistically model the \emph{default} behavior of the latent variables. The latter requirement can be particularly tedious for non-trivial latent variable models (\textit{e.g.} trees; \citealp{Corro2019DifferentiableAutoencoder}).  

\subsection{Resulting Objective}
Applying both of the previous simplifications to the semi-supervised objective in Eq. \ref{JALPHA} leads to the following objective:

 \begin{align}
 & \sum_{(x, y) \in L}\Bigr( \log p_\theta(x|y)+ \alpha \hspace{1mm} \log q_\phi(y|x)\Bigl)\nonumber\\
  &+ \sum_{x \in U} \mathbb{E}_{y\sim q_\phi(y|x)}\left[\log p_\theta(x|y)\right]) \label{JALPHAMOD}
 \end{align}
 As can be seen, the first ELBo in Eq. \ref{JALPHA} turns into a supervised conditional generation objective, while the second ELBo turns into a reconstruction term that relies only on $y$. Nevertheless, we stress that the second term is still an ELBo, and the whole objective is still a VAE-based semi-supervised learning objective. 
  It should also be noted that, without $z$, the latent variables cannot provide the decoder with  the full information about a sentence and, therefore, cannot reach a state where each sample is \emph{reconstructed}. To avoid confusion, instead of \emph{reconstructing} from $y$, the role of the reconstruction term is better read in our case as \emph{raising the probability} of the sample at hand under the associated label $y$.

% In other words, to generate samples with the generative model, one has to use latent variables from the output of the encoder (given real input samples) instead of sampling from a known prior distribution. 

\section{Experiments}
In this section, we display comparisons between instances of standard SSVAEs and the same SSVAEs after applying the changes we propose.
\subsection{Setup}
\paragraph{Datasets} We consider 4 Datasets for our study: the IMDB \cite{Maas2011LearningAnalysis} and Yelp review \cite{li-etal-2018-delete} binary sentiment analysis datasets, and the AG News and DBPedia \cite{zhang2015character} topic classification datasets. The Datasets have been chosen to represent a range over different tasks (Sentiment Analysis and Topic Classification), different numbers of classes, and different sentence lengths.  A summary of dataset statistics is in Table \ref{tab:Data}. 

\begin{table}[!htbp]
    \centering
    \resizebox{0.5\textwidth}{!}{
    \begin{tabular}{c|c c c c c c c }
    \hline
    dataset & Labels & Av. Sample length &  N° Classes\\
    \hline \hline
    AG News & Topic  & 37.85$\pm$10.09 & 4\\
    DBPedia& Topic & 46.13$\pm$22.46 & 14\\
    IMDB & Sentiment & 233.79$\pm$173.72& 2 \\
    Yelp & Sentiment & 8.88$\pm$3.64 & 2\\
    %UD 2 & PoS-Tagging & 16.31$\pm$12.40 & 10K& 4K& 1K % 12.5K& 11.6K&  2.9K
    %& 2K& 17 \\
    %test set sizes: 25, 4.6, 1, 70
    \hline
    \end{tabular}}
    \caption{Dataset properties.}
    \label{tab:Data}
  \end{table}
As was done in \citet{Chen2020MixText:Classification}, we measure performance on the different datasets with equal numbers of samples. Accordingly, for each dataset, we randomly subsample 10K samples from the original training set as \emph{unlabeled} data. We also use 4K labeled samples a training set and 1K as development set. We use the original test sets from each dataset. All the samples are tokenized using a simple whitespace tokenizer. 

\paragraph{Network Architecture}
 The size of $z$ is set to 32. For experiments without $z$, we simply drop all the components associated to it from the network.
 
 The encoder consists of a pre-trained $300$-dimensional fastText \cite{Bojanowski2016} embedding layer, and 2 Bidirectional LSTM networks with $100$ hidden states each, one for each of the latent variables $y$ and $z$. The logits of $y$ are then obtained by passing the last state of its Bidirectional LSTM through a linear layer. Similarly the last state of the Bidirectional LSTM for $z$ is passed through a linear layer to obtain its mean parameter, and a linear layer with a softplus activation to obtain its standard deviation parameter.
 
As for the decoding step, to allow backpropagation, $z$ is sampled using the reparameterization trick \cite{Kingma2014Auto-encodingBayes}, and $y$ is sampled using the Gumbel-Softmax trick \cite{Jang2017CategoricalGumbel-softmax}. \citet{Xu2017VariationalClassification} have shown that latent variables are best exploited in SSVAEs when concatenated with the previous word at each generation step to obtain the next word. We design our decoder accordingly and use a 1-layered LSTM with size $200$. The only hyper-parameter we tune on the development set is $\alpha$, the coefficient weighting the supervised learning objective in Eq. \ref{JALPHA}, which is selected in the set $\{10^{0}, 10^{-1}, 10^{-2}, 10^{-3}\}$. Further implementation details are provided in Appendix \ref{HPARAMS}.

 \begin{table*}[!ht]
    \centering
    \resizebox{0.67\textwidth}{!}{%
    \begin{tabular}{c|c c c c }
    \hline
    Objective& AGNEWS & DBPedia& IMDB& Yelp \\
    \hline \hline
% Supervised& 81.02\textcolor{gray}{(0.64)}& 86.98\textcolor{gray}{(0.74)}& 92.47\textcolor{gray}{(0.48)}& 96.97\textcolor{gray}{(0.28)}\\ 
% SSVAE& 83.34\textcolor{gray}{(0.91)}& 87.89\textcolor{gray}{(0.54)}& 92.85\textcolor{gray}{(0.78)}& \textbf{97.75}\textcolor{gray}{(0.11)}\\ 
% SSVAE-\{KL\}& 83.87\textcolor{gray}{(0.47)}& \textbf{87.95}\textcolor{gray}{(0.19)}& 92.90\textcolor{gray}{(0.54)}& 97.58\textcolor{gray}{(0.13)}\\ 
% SSVAE-\{z\}& 81.90\textcolor{gray}{(5.17)}& 87.94\textcolor{gray}{(0.33)}& 93.60\textcolor{gray}{(0.74)}& 97.40\textcolor{gray}{(0.14)}\\ 
% SSVAE-\{KL, z\}& \textbf{84.79}\textcolor{gray}{(1.34)}& 87.85\textcolor{gray}{(0.29)}& \textbf{93.77}\textcolor{gray}{(0.61)}& 97.58\textcolor{gray}{(0.19)}\\ 
Supervised& 86.98\textcolor{gray}{(0.74)}& 96.97\textcolor{gray}{(0.28)}&81.02\textcolor{gray}{(0.64)}&  92.47\textcolor{gray}{(0.48)}\\ 
SSVAE&  87.89\textcolor{gray}{(0.54)}& \textbf{97.75}\textcolor{gray}{(0.11)}&83.34\textcolor{gray}{(0.91)}& 92.85\textcolor{gray}{(0.78)}\\ 
SSVAE-\{KL\}& \textbf{87.95}\textcolor{gray}{(0.19)}& 97.58\textcolor{gray}{(0.13)}& 83.87\textcolor{gray}{(0.47)}& 92.90\textcolor{gray}{(0.54)}\\ 
SSVAE-\{z\}&  87.94\textcolor{gray}{(0.33)}& 97.40\textcolor{gray}{(0.14)}&81.90\textcolor{gray}{(5.17)}& 93.60\textcolor{gray}{(0.74)}\\ 
SSVAE-\{KL, z\}&  87.85\textcolor{gray}{(0.29)}& 97.58\textcolor{gray}{(0.19)}&\textbf{84.79}\textcolor{gray}{(1.34)}& \textbf{93.77}\textcolor{gray}{(0.61)}\\ 

\hline
     \end{tabular}}
    \caption{Accuracies  On AGNEWS, DBPedia, IMDB, and Yelp. The values are averages over 5 runs with standard deviations between parentheses. The best score for each dataset and each amount of labeled data is given in bold.}
    \label{tab:resultsLabeled}
  \end{table*}
 
\subsection{Results}
\label{RESULTS}
\paragraph{Classification performance}
In Table \ref{tab:resultsLabeled}, we compare the performance of a standard SSVAE, to a SSVAE where we remove the KL-divergence (SSVAE-\{KL\}) another where $z$ is removed (SSVAE-\{z\}) and a third version where both the KL-divergence and $z$ are removed (SSVAE-\{KL, z\}). We measure performance on all datasets using \emph{accuracy}.
As a baseline, we also include the results of an objective that does not use unlabeled data. The architecture we use for this objective is simply the LSTM encoder that we use to obtain $y$ for the SSVAE objectives. This baseline is referred to as \emph{Supervised}. 
 
The aim of our experiment is to see whether we observe that there is a harm to the performance of SSVAEs when applying the proposed simplifications.
In Table~\ref{tab:resultsLabeled}, we see that applying both changes compares favorably to the standard SSVAE 2 times out of 4. The removal of $z$ yields the same comparison, while removing the $KL$ term causes improvement 3 times out of 4. For more extensive testing, we ran experiments for varying amounts of labeled data (from 1\% to 100\%; \textit{cf.} Appendix \ref{RESAPPEN}), and only found 4  statistically significant differences between SSVAE and its variants: 3 in favor of one of our Simplified SSVAEs, and 1 in favor of the standard SSVAE.
 
 We performed additional experiments in an out-of-domain setting (\textit{c.f} Appendix \ref{OOD}.) using our sentiment analysis datasets, and also observed improvements with our simplifications.

 \paragraph{Speeding Up the Learning Process} By removing the KL-divergence and the components associated with $z$, an improvement on the speed of the learning process is to be expected. This improvement is highly dependent on the model and on the implementation at hand. As an example, we measure the average speed of an optimization iteration for each dataset, and each version of SSVAE. In Table \ref{tab:speed}, the speed of each objective is displayed proportionally to the speed of standard SSVAEs.
 The calculations associated with the KL-divergence do not seem to slow down the iterations. However, removing $z$ and its associated components consistently cuts out a considerable proportion of the duration of optimization steps. This proportion ranges from 14\% (DBPedia) to 26\%(AGNEWS).% for our experiments. 
  \begin{table}[!h]
    \centering
    \resizebox{0.47\textwidth}{!}{%
    \begin{tabular}{c|| c c c }
    \hline
    Dataset& SSVAE-\{KL\} &SSVAE-\{z\} &SSVAE-\{KL, z\}\\
    \hline \hline
    % IMDB& 0.316\textcolor{gray}{(0.08)}& 0.322\textcolor{gray}{(0.08)}& 0.260\textcolor{gray}{(0.08)}& \textbf{0.258}\textcolor{gray}{(0.08)}\\ 
    % AGNEWS& 0.124\textcolor{gray}{(0.12)}& 0.113\textcolor{gray}{(0.09)}& \textbf{0.092}\textcolor{gray}{(0.08)}& \textbf{0.092}\textcolor{gray}{(0.10)} \\ 
    % Yelp& 0.072\textcolor{gray}{(0.05)}& 0.071\textcolor{gray}{(0.11)}& \textbf{0.059}\textcolor{gray}{(0.10)} & \textbf{0.059}\textcolor{gray}{(0.11)}\\ 
    % DBPedia& 0.180\textcolor{gray}{(0.09)}& 0.185\textcolor{gray}{(0.10)}& \textbf{0.155}\textcolor{gray}{(0.11)}& 0.156\textcolor{gray}{(0.10)} \\
    
    AGNEWS& %1.0\textcolor{gray}{(0.967)}&
    0.911\textcolor{gray}{(0.73)}& \textbf{0.742}\textcolor{gray}{(0.65)}& \textbf{0.742}\textcolor{gray}{(0.81)} \\ 
    DBPedia& %1.0\textcolor{gray}{(0.5)}& 
    1.03\textcolor{gray}{(0.56)}& \textbf{0.861}\textcolor{gray}{(0.61)}& 0.867\textcolor{gray}{(0.56)} \\
    IMDB& %1.0\textcolor{gray}{(0.25)}&
    1.018\textcolor{gray}{(0.25)}& 0.822\textcolor{gray}{(0.25)}& \textbf{0.816}\textcolor{gray}{(0.25)}\\ 
    
    Yelp& %1.0\textcolor{gray}{(0.69)}&
    0.986\textcolor{gray}{(1.52)}& \textbf{0.819}\textcolor{gray}{(1.39)} & \textbf{0.819}\textcolor{gray}{(1.52)}\\

   %UD& 0.149\textcolor{gray}{(0.07)}& 0.150\textcolor{gray}{(0.09)}& 0.126\textcolor{gray}{(0.09)}& \textbf{0.12}\textcolor{gray}{(0.06)}\\ 
    \hline
     \end{tabular}}
    \caption{Training durations for each objective relative to standard SSVAE, averaged over 200 iterations. Standard deviations are given between parentheses. Lowest duration for each dataset is given in bold.}
    %significance is confidence interval * 0.1375
    \label{tab:speed}
  \end{table}
%  \paragraph{Out-of-Domain Representation Transfer}
%  In a similar spirit, we use the latent variable $z$ of a SSVAE, trained for sentiment analysis on one domain,  as in input to a linear classifier that will be trained for the same task on the other domain. This is to see whether simplifying SSVAEs by removing the KL-divergence hurts the representations in $z$, and diminishes transfer performance. The results are in Table \ref{tab:transfer}.
 
% \begin{table}[!htbp]
%     \centering
%     \resizebox{0.45\textwidth}{!}{%
%     \begin{tabular}{c||c c}
%     \hline
%     Dataset&  SSVAE& SSVAE-\{KL\} \\
%     \hline \hline
%     IMDB$\longrightarrow$Yelp
%     & \textbf{81.42}\textcolor{gray}{(2.25)}
%     & 81.02\textcolor{gray}{(2.64)}
%     \\ 
    
%     Yelp$\longrightarrow$IMDB 
%     & \textbf{71.37}\textcolor{gray}{(1.26)}
%     & 71.03\textcolor{gray}{(0.44)}
%     \\ 
%     \hline
%      \end{tabular}}
%     \caption{Representation transfer performance between IMDB and Yelp with or without the Kullback-Leibler term. The best objective for each transfer direction is given in bold.}
%     %significance is confidence interval * 0.1375
%     \label{tab:transfer}
%   \end{table}

%  Although the standard SSVAE objective takes the lead in both transfer directions, the difference is well within the standard deviation range, and therefore shows no significance.
\section{Related Works}
\label{RELATED}
After the pioneering work of \citet{Kingma2014a}, SSVAEs
were extended to tasks such as morphological inflections~\cite{Wolf-Sonkin2018AInflection}, controllable speech synthesis~\cite{Habib2019Semi-SupervisedSynthesis}, parsing~\cite{Corro2019DifferentiableAutoencoder}, sequential labeling~\cite{Chen2018VariationalLearning} among many others.
VAE internals have also been \emph{tweaked} in various manners to improve the learning performance.
For instance, \citet{Gururangan2020VariationalClassification} introduce a low resource pretraining scheme to improve transfer with VAEs, while \citet{Zhang2019AdvancesInference} propose to use the deterministic ancestor of a latent
variable to perform classification, and constrain it with an adversarial term to have it abide by the values of the random latent variable.

 While our work is a focused contribution dedicated to the theoretical soundness and the practical advantages of two simplifications to the SSVAE framework for text classifications, it could be extendend to other tasks involving text generation as the unsupervised VAE objective. For instance, the work of \citet{Corro2019DifferentiableAutoencoder} shows that semi-supervised dependency parsing scores higher with both the changes we study.
% As such, researchers pursuing this direction should be aware that an objective that focuses the conditional generation learning signal on $y$ (as if $z$ was removed), can be formulated for a SSVAE that uses both $y$ and $z$, and thus preserves the conditional reconstruction ability of the SSVAE. The derivation of such an objective is based on a partial use of the IWAE objective \cite{Burda2016ImportanceAutoencoders}, and is detailed in Appendix \ref{PIWOAPPEN}.

\section{Conclusion}
Starting from the observation that SSVAEs can be viewed as the combination of a supervised learning signal with an unsupervised conditional generation learning signal, we show that this framework needs neither to include a KL-divergence nor an unobserved latent variable ($z$) when dealing with text classification. We subsequently perform experimental comparisons between standard SSVAEs and simplified SSVAEs that indicate that they are globally equivalent in performance.

Our changes provide a number of practical advantages. First, removing the KL-divergence frees practitioners from choosing priors for the variables they use, and allows information to flow freely into these variables. Second, removing the latent variable $z$ from the computational graph speeds up computation and shrinks the size of the network. Despite their popularity, VAEs are often tedious to train for NLP tasks. In that regard, our simplifications should facilite their usage in future works.

 \section*{Acknowledgments}
This work is supported by the PARSITI project grant (ANR-16-CE33-0021) given by the French National Research Agency (ANR), the \emph{Laboratoire d’excellence “Empirical Foundations of Linguistics”} (ANR-10-LABX-0083), as well as the ONTORULE project. It was also granted access to the HPC resources of IDRIS under the allocation 20XX-AD011012112 made by GENCI.
\bibliography{references, bib2}
\bibliographystyle{acl_natbib}

\clearpage
\appendix
\section{Implementation Details}
\label{HPARAMS}
\paragraph{Training and validation data splits}
We sample 5 \emph{labeled} data splits of size 1K. Each of these 5 splits will, in turn, play the role of validation set for one experiment, while the other 4 splits are used for training. Looping over these splits yields 5 runs for each experiment. The results we display are the average (and standard deviation) of the results for each of these runs. The validation score serves selecting hyper-parameters (in our case only $\alpha$ from Eq. \ref{JALPHA}). The final test scores are measured on the original test set of each dataset. 
\paragraph{Probabilistic Graphical Model}
For models that use both $z$ and $y$, we consider the latent variables to be conditionally independent in the inference model (\textit{i.e.} $q_\phi(y, z|x)=q_\phi(y|x)q_\phi(z|x)$) ) and independent in the generation model (\textit{i.e} $p_\theta(y, z)= p(y)p(z)$). 
%The generative model for models using both $y$ and $z$ is $p_\theta(x)= \prod_i p(x_i|x_{<i}, y, z)$. 
%For sequence-level tasks, the observed variable is a sentence $x$. However, for the word-level task (PoS Tagging), the observed variable is a word $x_i$ that we generate conditioned on the corresponding latent variables $z_i$ (when available) and $y_i$ and the previous words $x_{<i}$. Therefore, the generative model for word-level tasks for models using both $y$ and $z$ is $p_\theta(x)= \prod_i p(x_i|x_{<i}, y_i, z_i)$ while for sentence-level tasks it is $p_\theta(x)= \prod_i p(x_i|x_{<i}, y, z)$. 

\begin{table*}[!ht]
    \centering
    \resizebox{0.9\textwidth}{!}{%
    \begin{tabular}{c||c c c c c}
    \hline
    Dataset&  Supervised& SSVAE& SSVAE-\{KL\} &SSVAE-\{z\} &SSVAE-\{KL, z\}\\
    \hline \hline
    IMDB$\longrightarrow$Yelp
    & 59.07\textcolor{gray}{(1.19)}
    & 61.78\textcolor{gray}{(6.03)}
    & 68.67$^+$\textcolor{gray}{(4.85)}
    & \textbf{71.30}$^+$\textcolor{gray}{(7.67)}
    & 64.69$^+$\textcolor{gray}{(3.84)}
    \\ 
    
    Yelp$\longrightarrow$IMDB 
    & 66.17\textcolor{gray}{(2.62)}
    & \textbf{69.54}\textcolor{gray}{(2.49)}
    & 66.67\textcolor{gray}{(3.26)}
    & 65.15\textcolor{gray}{(2.31)}
    & 66.13\textcolor{gray}{(3.82)}
    \\ 
    \hline
     \end{tabular}}
    \caption{Out-of-domain Accuracies between IMDB and Yelp for the different objectives. The best objective for each out-of-domain inference direction is given in bold. The scores displaying statistically significant improvement compared to the score of the supervised objective are marked with $^+$  }
    %significance is confidence interval * 0.1375
    \label{tab:OOD}
  \end{table*}
\paragraph{Training Procedure}
 We use the STL estimator \cite{Roeder2017StickingInference} which is a low-variance unbiased gradient estimator for ELBo. 
 
 The network is optimized using ADAM \cite{Kingma2015}, with a learning rate of 4e-3 and a dropout rate of 0.5. If the accuracy on the validation set doesn't increase for 4 epochs, the learning rate is divided by 4. If it doesn't increase for 8 epochs, the training is stopped. For objectives that include a KL-divergence, we scale it with a coefficient that is null for 3K steps then linearly increased to 1 for the following 3K steps to avoid posterior collapse \cite{Li2020AText}. 
 \section{Out-of-domain experiments}
 \label{OOD}
 The sentiment analysis tasks we use for these experiments take place in
 different domains (Restaurant reviews for Yelp, and Movie reviews for IMDB). 
 Using models trained for each domain (with \%100 of the data), we measure performance on the other domain to see whether the changes we study have an effect on out-of-domain generalization. In Table \ref{tab:OOD}, we compare the out-of-domain performances of each of the objectives to that of the baseline that doesn't use unlabeled data (\emph{Supervised}). 
 
 The table shows no statistically significant gains from using unlabeled Yelp training data for inference on IMDB. This is to be expected as reviews from Yelp are drastically shorter than those from IMDB (\textit{cf.} Table \ref{tab:Data}). However, for out-of-domain inference in the opposite direction, all the semi-supervised objectives except the standard SSVAE show statistically significant gains. Removing the KL-divergence to accumulate more information in $y$, and removing $z$ to have conditional generation exclusively rely on $y$ seem to be effective to help generalization beyond the original domain of the task. 
 
\section{Results Over Varying Amounts of Data}
\label{RESAPPEN}
We display results with varying amounts of data in Table \ref{tab:resultsLabeledAppen}.

\begin{table*}[!ht]
    \centering
    \resizebox{0.9\textwidth}{!}{%
    \begin{tabular}{c c|c c c c c }
    \hline
    Dataset&  Objective & 1\% &  3\% & 10\% & 30\% & 100\% \\
    \hline \hline
    \multirow{5}{*}{IMDB}
&Supervised& 54.62\textcolor{gray}{(3.30)}& 56.47\textcolor{gray}{(1.02)}& 62.01\textcolor{gray}{(2.75)}& 69.65\textcolor{gray}{(2.02)}& 81.02\textcolor{gray}{(0.64)}\\ 
&SSVAE& 53.92\textcolor{gray}{(2.34)}& 56.03\textcolor{gray}{(4.20)}& 62.15\textcolor{gray}{(5.03)}& 75.39\textcolor{gray}{(0.49)}& 83.34\textcolor{gray}{(0.91)}\\ 
&SSVAE-\{KL\}& 52.70\textcolor{gray}{(1.72)}& 54.95\textcolor{gray}{(0.77)}& 62.37\textcolor{gray}{(4.45)}& 74.18\textcolor{gray}{(1.97)}& 83.87\textcolor{gray}{(0.47)}\\ 
&SSVAE-\{z\}& \textbf{54.15}\textcolor{gray}{(2.46)}& \textbf{56.86}\textcolor{gray}{(1.77)}& 62.15\textcolor{gray}{(2.87)}& 75.42\textcolor{gray}{(1.80)}& 81.90\textcolor{gray}{(5.17)}\\ 
&SSVAE-\{KL, z\}& 53.51\textcolor{gray}{(1.99)}& 56.58\textcolor{gray}{(2.22)}& \textbf{63.24}\textcolor{gray}{(4.15)}& \textbf{75.87}\textcolor{gray}{(1.30)}& \textbf{84.79}\textcolor{gray}{(1.34)}\\ 
\hline
    \hline
    \multirow{5}{*}{AGNEWS}
&Supervised& 68.60\textcolor{gray}{(4.88)}& 75.92\textcolor{gray}{(1.74)}& 81.96\textcolor{gray}{(0.83)}& 84.59\textcolor{gray}{(0.67)}& 86.98\textcolor{gray}{(0.74)}\\ 
&SSVAE& 65.79\textcolor{gray}{(5.02)}& 75.95\textcolor{gray}{(1.27)}& 82.47\textcolor{gray}{(0.43)}& 85.50\textcolor{gray}{(0.30)}& 87.89\textcolor{gray}{(0.54)}\\ 
&SSVAE-\{KL\}& \textbf{68.56}\textcolor{gray}{(1.89)}& 76.25\textcolor{gray}{(2.21)}& 82.76\textcolor{gray}{(0.45)}& 85.73\textcolor{gray}{(0.80)}& \textbf{87.95}\textcolor{gray}{(0.19)}\\ 
&SSVAE-\{z\}& 67.13\textcolor{gray}{(6.55)}& \textbf{77.28}\textcolor{gray}{(1.81)}& \textbf{83.48}$^*$\textcolor{gray}{(0.75)}& \textbf{85.75}\textcolor{gray}{(0.74)}& 87.94\textcolor{gray}{(0.33)}\\ 
&SSVAE-\{KL, z\}& 66.96\textcolor{gray}{(3.42)}& 76.47\textcolor{gray}{(1.24)}& 82.58\textcolor{gray}{(0.97)}& 85.51\textcolor{gray}{(0.57)}& 87.85\textcolor{gray}{(0.29)}\\ 
\hline
    \hline
    \multirow{5}{*}{Yelp}
&Supervised& 70.32\textcolor{gray}{(1.84)}& 76.32\textcolor{gray}{(2.07)}& 83.41\textcolor{gray}{(1.75)}& 87.85\textcolor{gray}{(0.58)}& 92.47\textcolor{gray}{(0.48)}\\ 
&SSVAE& \textbf{71.34}\textcolor{gray}{(1.93)}& 76.96\textcolor{gray}{(1.64)}& 82.96\textcolor{gray}{(0.69)}& 89.35\textcolor{gray}{(0.39)}& 92.85\textcolor{gray}{(0.78)}\\ 
&SSVAE-\{KL\}& 69.85\textcolor{gray}{(2.86)}& 76.82\textcolor{gray}{(1.31)}& 82.90\textcolor{gray}{(2.23)}& 88.33\textcolor{gray}{(0.99)}& 92.90\textcolor{gray}{(0.54)}\\ 
&SSVAE-\{z\}& 68.74\textcolor{gray}{(2.95)}& \textbf{78.26}\textcolor{gray}{(1.70)}& 84.11\textcolor{gray}{(1.25)}& \textbf{90.27}$^*$\textcolor{gray}{(0.28)}& 93.60\textcolor{gray}{(0.74)}\\ 
&SSVAE-\{KL, z\}& 69.21\textcolor{gray}{(1.10)}& 77.30\textcolor{gray}{(2.57)}& \textbf{85.02}$^*$\textcolor{gray}{(1.24)}& 89.74\textcolor{gray}{(1.31)}& \textbf{93.77}\textcolor{gray}{(0.61)}\\ 
% \hline
%     \hline
%     \multirow{5}{*}{UD}
% &Supervised& 73.65\textcolor{gray}{(0.84)}& 81.99\textcolor{gray}{(0.18)}& 86.74\textcolor{gray}{(0.36)}& 89.32\textcolor{gray}{(0.07)}& 91.35\textcolor{gray}{(0.20)}\\ 
% &SSVAE& 72.85\textcolor{gray}{(0.63)}& 81.69\textcolor{gray}{(0.22)}& 86.58\textcolor{gray}{(0.30)}& 89.27\textcolor{gray}{(0.30)}& 91.35\textcolor{gray}{(0.13)}\\ 
% &SSVAE-\{KL\}& 73.22\textcolor{gray}{(0.78)}& 81.73\textcolor{gray}{(0.24)}& 86.60\textcolor{gray}{(0.32)}& 89.10\textcolor{gray}{(0.32)}& \textbf{91.40}\textcolor{gray}{(0.05)}\\ 
% &SSVAE-\{z\}& \textbf{73.35}\textcolor{gray}{(0.65)}& \textbf{81.96}\textcolor{gray}{(0.40)}& \textbf{86.61}\textcolor{gray}{(0.31)}& 89.27\textcolor{gray}{(0.33)}& 91.35\textcolor{gray}{(0.20)}\\ 
% &SSVAE-\{KL, z\}& 73.08\textcolor{gray}{(0.68)}& 81.74\textcolor{gray}{(0.27)}& 86.27\textcolor{gray}{(0.51)}& \textbf{89.38}\textcolor{gray}{(0.33)}& \textbf{91.40}\textcolor{gray}{(0.06)}\\ 

\hline
    \hline
    \multirow{5}{*}{DBPedia}
&Supervised& 63.67\textcolor{gray}{(1.74)}& 81.49\textcolor{gray}{(2.25)}& 90.56\textcolor{gray}{(1.21)}& 94.63\textcolor{gray}{(0.32)}& 96.97\textcolor{gray}{(0.28)}\\ 
&SSVAE& 64.42\textcolor{gray}{(1.83)}& 83.16\textcolor{gray}{(1.49)}& 92.95\textcolor{gray}{(0.82)}& 96.26\textcolor{gray}{(0.25)}& \textbf{97.75}\textcolor{gray}{(0.11)}\\ 
&SSVAE-\{KL\}& \textbf{66.09}\textcolor{gray}{(3.05)}& 81.97\textcolor{gray}{(1.54)}& \textbf{93.64}\textcolor{gray}{(0.76)}& 96.32\textcolor{gray}{(0.28)}& 97.58\textcolor{gray}{(0.13)}\\ 
&SSVAE-\{z\}& 62.56\textcolor{gray}{(5.60)}& \textbf{83.40}\textcolor{gray}{(2.42)}& 93.37\textcolor{gray}{(1.00)}& \textbf{96.39}\textcolor{gray}{(0.21)}& 97.40$^\dag$\textcolor{gray}{(0.14)}\\ 
&SSVAE-\{KL, z\}& 62.15\textcolor{gray}{(1.68)}& 82.67\textcolor{gray}{(2.16)}& 93.40\textcolor{gray}{(1.10)}& 96.31\textcolor{gray}{(0.24)}& 97.58\textcolor{gray}{(0.19)}\\ 

\hline
     \end{tabular}}
    \caption{Accuracies on IMDB, AGNEWS, Yelp and DBPedia with varying amount of \textbf{labeled} data. The values are averages over 5 runs with standard deviations between parentheses. The best score for each dataset and each amount of labeled data is given in bold. Each semi-supervised objective that scores above (resp. below) SSVAE with p-value<0.05 is marked with $^*$ (resp. $^\dag$)}
    \label{tab:resultsLabeledAppen}
  \end{table*}

\end{document}